%% file: main.tex
\documentclass[11pt, a4paper, logo, onecolumn,copyright]{googledeepmind}

\usepackage[numbers, sort&compress, square]{natbib}
\title{Gemini Embedding 2: A Native Multimodal Embedding Model from Gemini}

\usepackage[utf8]{inputenc} %
\usepackage[T1]{fontenc}    %
\usepackage{hyperref}       %
\usepackage{url}            %
\usepackage{booktabs}       %
\usepackage{nicefrac}       %
\usepackage{microtype}      %
\usepackage{amsmath}
\usepackage{graphicx}
\usepackage{multicol}
\usepackage{siunitx}
\usepackage{array}
\usepackage[nameinlink]{cleveref}
\usepackage{bbm}
\usepackage{multirow}
\usepackage{subfig}
\usepackage{soul}
\usepackage{floatrow}
\usepackage{float}
\usepackage{wrapfig}
\usepackage{blindtext}
\usepackage{tablefootnote}
\usepackage{amsfonts}
\usepackage[flushleft]{threeparttable}
\usepackage{colortbl}
\usepackage{bbding}
\usepackage{lipsum}
\usepackage{svg}
\usepackage{booktabs}
\usepackage{multirow}
\usepackage{graphicx} %

\usepackage[table]{xcolor}
\usepackage{collcell}
\usepackage{xfp} 
\usepackage{booktabs}
\usepackage{multirow}
\usepackage{graphicx}

\newcommand{\OursTwo}{Gemini Embedding 2}

\input{commands}

\author[*]{Madhuri Shanbhogue}
\author[*]{Zhe Li}
\author[*]{Shanfeng Zhang}
\author[*]{Gustavo Hern{\'{a}}ndez {\'{A}}brego}
\author[*]{Shih-Cheng Huang}
\author[*]{Aashi Jain}
\author[ \hspace{-0.2em}]{Daniel Salz}
\author[ \hspace{-0.2em}]{Sonam Goenka}
\author[ \hspace{-0.2em}]{Chaitra Hegde}
\author[ \hspace{-0.2em}]{Ji Ma}
\author[ \hspace{-0.2em}]{Feiyang Chen}
\author[ \hspace{-0.2em}]{Jiaxing Wu}
\author[ \hspace{-0.2em}]{Tanmaya Dabral}
\author[ \hspace{-0.2em}]{Babak Samari}
\author[ \hspace{-0.2em}]{Kevin Poulet}
\author[ \hspace{-0.2em}]{Daniel Cer}
\author[ \hspace{-0.2em}]{Kaifeng Chen}
\author[ \hspace{-0.2em}]{Paul Suganathan}
\author[ \hspace{-0.2em}]{Hui Hui}
\author[ \hspace{-0.2em}]{Jovan Andonov}
\author[ \hspace{-0.2em}]{Philippe Schlattner}
\author[ \hspace{-0.2em}]{Jay Han}
\author[ \hspace{-0.2em}]{Iftekhar Naim}
\author[ \hspace{-0.2em}]{Wing Lowe}
\author[ \hspace{-0.2em}]{Vladimir Pchelin}
\author[ \hspace{-0.2em}]{Albert Yang}
\author[ \hspace{-0.2em}]{Yi-Ting Chen}
\author[ \hspace{-0.2em}]{Zhongli Ding}
\author[ \hspace{-0.2em}]{Grace Zhang}
\author[ \hspace{-0.2em}]{Georg Heigold}
\author[ \hspace{-0.2em}]{Yichang Chen}
\author[ \hspace{-0.2em}]{Antoine Reveillon}
\author[ \hspace{-0.2em}]{Brendan Mccloskey}
\author[ \hspace{-0.2em}]{Wenlei Zhou}
\author[ \hspace{-0.2em}]{Dahun Kim}
\author[ \hspace{-0.2em}]{Rui Meng}
\author[ \hspace{-0.2em}]{Emma Wang}
\author[ \hspace{-0.2em}]{Jack Zheng}
\author[ \hspace{-0.2em}]{Halley Fede}
\author[ \hspace{-0.2em}]{Zhen Yang}
\author[ \hspace{-0.2em}]{Keegan Mosley}
\author[ \hspace{-0.2em}]{Brian Potetz}
\author[ \hspace{-0.2em}]{Sahil Dua}
\author[ \hspace{-0.2em}]{Henrique Schechter Vera}
\author[ \hspace{-0.2em}]{Shen Gao}
\author[ \hspace{-0.2em}]{Hesen Zhang}
\author[ \hspace{-0.2em}]{Andreas Hess}
\author[ \hspace{-0.2em}]{Hengxuan Ying}
\author[ \hspace{-0.2em}]{Alberto Montes}
\author[ \hspace{-0.2em}]{Karan Gill}
\author[ \hspace{-0.2em}]{Min Choi}
\author[ \hspace{-0.2em}]{Sebastian Russo}
\author[ \hspace{-0.2em}]{Anja Hauth}
\author[ \hspace{-0.2em}]{Jinhyuk Lee}
\author[ \hspace{-0.2em}]{Michael Boratko}
\author[ \hspace{-0.2em}]{Megan Barnes}
\author[ \hspace{-0.2em}]{Vikram Rao}
\author[ \hspace{-0.2em}]{Claudiu Musat}
\author[ \hspace{-0.2em}]{Cyril Allauzen}
\author[ \hspace{-0.2em}]{Ehsan Variani}
\author[ \hspace{-0.2em}]{Shankar Kumar}
\author[ \hspace{-0.2em}]{Tom Bagby}
\author[ \hspace{-0.2em}]{Junyi Jiao}
\author[ \hspace{-0.2em}]{Yang Gu}
\author[ \hspace{-0.2em}]{Tengxin Li}
\author[ \hspace{-0.2em}]{Ayush Agrawal}
\author[ \hspace{-0.2em}]{Roberto Santana}
\author[ \hspace{-0.2em}]{Dev Nath}
\author[ \hspace{-0.2em}]{Stephen Karukas}
\author[ \hspace{-0.2em}]{Shuoxuan Han}
\author[ \hspace{-0.2em}]{Lucia Loher}
\author[ \hspace{-0.2em}]{Alice Twu}
\author[ \hspace{-0.2em}]{Nidhi Vyas}
\author[ \hspace{-0.2em}]{Siddharth Bhai}
\author[ \hspace{-0.2em}]{Frank Palma Gomez}
\author[ \hspace{-0.2em}]{Wangyuan Zhang}
\author[ \hspace{-0.2em}]{Chaoren Liu}
\author[ \hspace{-0.2em}]{Jizheng Yang}
\author[ \hspace{-0.2em}]{Steve Qiu}
\author[ \hspace{-0.2em}]{Shijie Zhang}
\author[ \hspace{-0.2em}]{Sujay Kulkarni}
\author[ \hspace{-0.2em}]{Sascha Rothe}
\author[ \hspace{-0.2em}]{Sean Nakamoto}
\author[ \hspace{-0.2em}]{Raphael Hoffmann}
\author[ \hspace{-0.2em}]{Zach Gleicher}
\author[ \hspace{-0.2em}]{Yunhsuan Sung}
\author[ \hspace{-0.2em}]{Qin Yin}
\author[ \hspace{-0.2em}]{Tom Duerig}
\author[ \hspace{-0.2em}]{Mojtaba Seyedhosseini}

\affil[ \hspace{-0.2em}]{Gemini Embedding Team, Google\footnote{See Contributions and Acknowledgments section. $^*$Equal contributions.}}

\begin{abstract}

We introduce \OursTwo , a native multimodal embedding model that allows embedding video, audio, image, and text modalities in a unified representation space. We leverage the \omniModal\ capabilities of Gemini to produce embeddings for arbitrary combinations of interleaved inputs across all these modalities that generalize well across a wide variety of tasks. Applying large-scale contrastive learning in a multi-task multi-stage training setup, we achieve state-of-the-art performance on key embedding benchmarks including unimodal, cross-modal, and multimodal retrieval spanning a diverse set of tasks. We show that our embedding model demonstrates strong performance (with a score of 62.9 R@1 on MSCOCO, 68.8 NDCG@10 on Vatex, 69.9 on MTEB multilingual and 84.0 on MTEB Code) across a variety of tasks surpassing the performance of specialized models. These unified capabilities make \OursTwo\ a promising candidate for downstream use cases such as RAG, recommendation and search. Furthermore, its robust zero-shot performance across distinct fields -- from astronomy and bioscience to fine arts and the culinary arts -- establishes it as a highly reliable, out-of-the-box representation even for specialized domains.
\
\end{abstract}

\begin{document}

\maketitle

\input{sections/introduction}
\input{sections/related}
\input{sections/modeling}

\input{sections/evaluation}

\input{sections/ablation}

\input{sections/future}
\input{sections/conclusion}

\bibliography{main}

\clearpage
\input{sections/appendix}

\clearpage
\input{sections/contributions}

\end{document}

%% file: commands.tex
\usepackage{xspace}
\newcommand{\Ours}{Gemini Embedding}
\newcommand{\ours}{Gemini Embedding}
\newcommand{\Ourstwo}{Gemini Embedding 2}
\newcommand{\ourstwo}{Gemini Embedding 2}
\newcommand{\amazonnova}{Amazon Nova MME}
\newcommand{\legacymme}{multimodalembedding@001}
\newcommand{\voyagemultimodal}{Voyage-3.5-multimodal}
\newcommand{\omnimodal}{multimodal}
\newcommand{\omniModal}{multimodal}

\newcommand{\beginOmniModal}{Multimodal}

\newfloatcommand{capbtabbox}{table}[][\FBwidth]

\newcommand{\eat}[1]{}
\definecolor{darkgreen}{RGB}{0, 102, 0}

\crefformat{section}{\S#2#1#3}

\newcommand{\pairsim}{\operatorname{sim}}
\newcommand{\meanpool}{\texttt{mean\_pool}}
\newcommand{\mask}{\texttt{mask}}

%% file: sections/introduction.tex
\section{Introduction}

Embedding models provide dense vector representations capturing semantic information that is crucial for adaptation in a wide range of downstream tasks. With foundational models being natively \omniModal\ and powered with exceptionally growing capabilities, it is important to ensure embedding models capture semantic information within and across all modalities in a coherent manner. Such general-purpose embedding models will also enhance the performance across a broad spectrum of applications like video recommendations and document search which are rich in information across different modalities but since the contained modalities are not inherently homogenous, they can benefit from having rich semantic information from across all modalities.

Existing \omniModal\ embedding models like CLIP \citep{radford2021learning}, ALIGN \citep{jia2021scaling}, SigLIP 2 \citep{tschannen2025siglip}, CoCa \citep{yu2022cocacontrastivecaptionersimagetext} embed heterogenous modalities by using paired cross-modal data and training modality-specific encoders to encode them into a unified vector space. This late-fusion approach results in good unimodal and cross-modal capabilities but has a key limitation in handling mixed-modality inputs and lacks richness since it does not utilize interactions between modalities. With advances in \beginOmniModal\ Large Language Models (MLLMs), it is now possible to achieve semantically richer embeddings enabled by the deep fusion of cross-modal interactions.

\input{figures/main_figure}
In this work, we introduce a generalizable \omniModal\ embedding model that embeds video, audio, image, text modalities, and any arbitrary combination thereof into a single representation space. The \omniModal\ \Ourstwo\ is trained by leveraging Gemini's~\citep{comanici2025gemini} capabilities and utilizing multi-task training with a diverse set of tasks resulting in a model that captures various interactions between modalities. \Cref{fig:main} shows a high-level representation of how \omniModal\ \Ourstwo\ maps the heterogenous sources into a unified vector space. The curated set of tasks help the model generalize across a wide variety of enterprise use cases like document retrieval, video recommendation, audio-based search, and RAG applications~\citep{lewis2020retrieval}. Crucially, enabling the model to handle interleaved sequences of images, text, and video facilitates complex, novel retrieval paradigms—such as zeroing in on specific temporal events in a video using combined visual and textual prompts. Using Gemini's capabilities we also show that native audio understanding and native \omniModal\ understanding outperforms text-based alternatives like ASR or captioning. 

\input{figures/spider}
We evaluate comprehensively on a wide variety of benchmarks, both academic-focused and enterprise-focused. As shown in \Cref{fig:spider}, our model achieves state-of-the-art performance compared to other models. For evaluating the text embedding capabilities, we rely on the Massive Multilingual Text Embedding Benchmark (MMTEB) \citep{enevoldsen2025mmteb} which consists of multi-lingual tasks spanning key downstream embedding use cases like retrieval, clustering, classification, etc. \Ourstwo\ achieves state-of-the-art performance on multilingual and code surpassing existing models on the leaderboard. We demonstrate strong numbers on a broad range of cross-modal retrieval benchmarks like MSCOCO \citep{chen2015microsoftcococaptionsdata}, Flickr30k \citep{plummer2016flickr30kentitiescollectingregiontophrase}, and MSR-VTT \citep{Xu2016MSRVTTAL}. We also demonstrate the model's ability to generalize to most multimodal retrieval tasks in general as well as specialized domains. 

%% file: figures/main_figure.tex
\begin{figure}[t]
\centering
\includegraphics[width=1\textwidth]{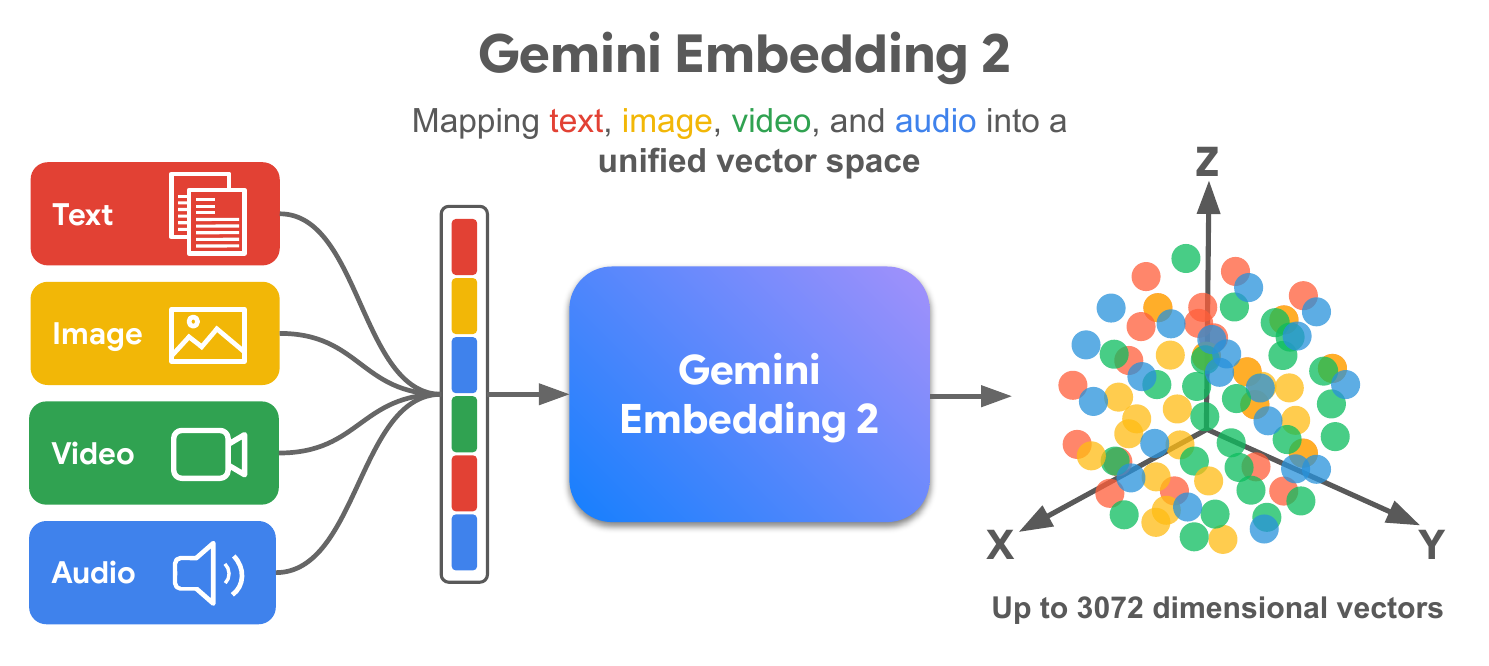}
\caption{Conceptual overview of the Gemini Embedding 2 workflow. The model natively processes heterogeneous inputs—text, images, video, audio, documents, and their combinations—mapping them into a single, unified high-dimensional vector space where cross-modal semantic relationships are preserved.
}
\label{fig:main}
\end{figure}

%% file: figures/spider.tex
\begin{figure}[t]
\centering
\includegraphics[width=0.85\textwidth]{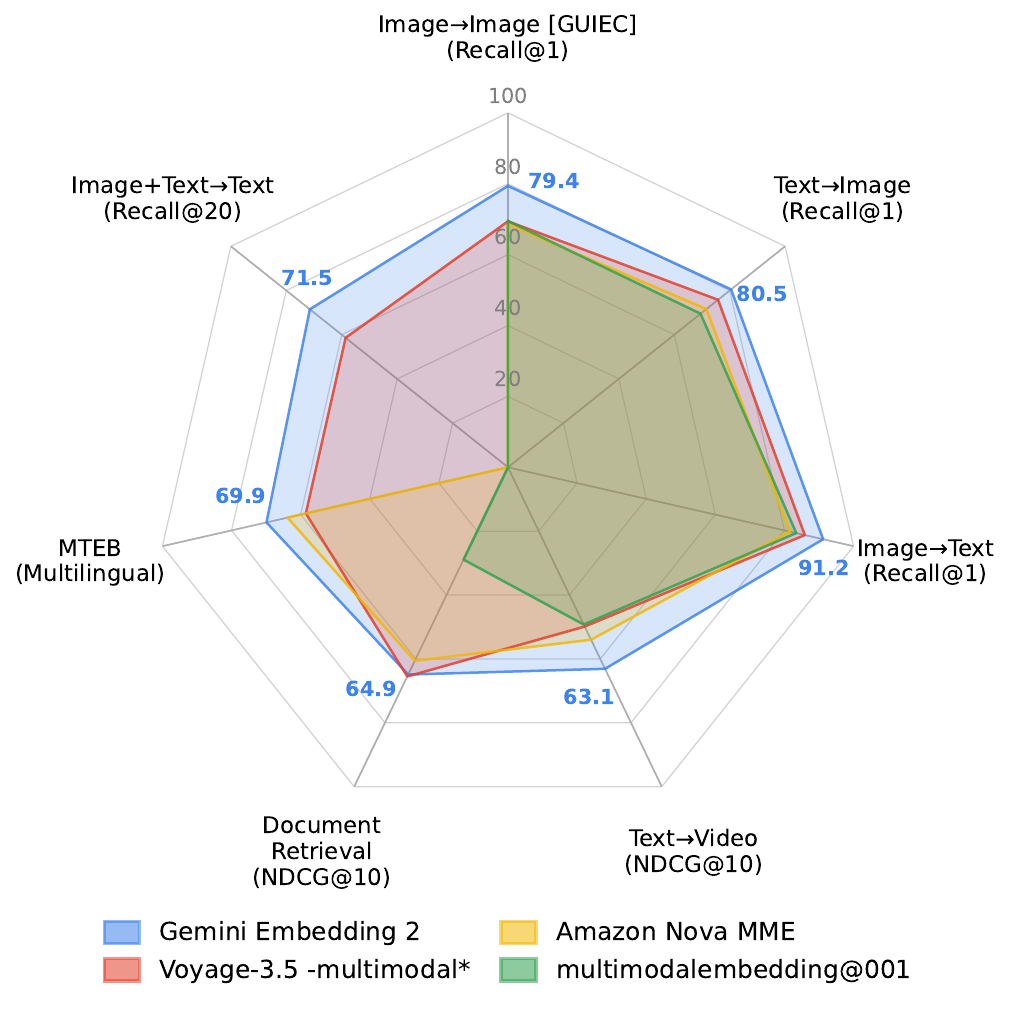}
\caption{Gemini Embedding 2 shows strong performance across multimodal retrieval tasks spanning image, text, video, and document modalities. \\ $^*$\text{MTEB number is reported for Voyage-3.5 since Voyage-3.5-multimodal does not report MTEB}.}
\label{fig:spider}
\end{figure}

%% file: sections/related.tex
\vspace{-0.5em}
\section{Related Work}

\paragraph{Large Language Models as Text Embedders} The paradigm of text embedding models has matured from relying on purely encoder-only architectures (e.g., BERT \citep{devlin2019bert}, RoBERTa \citep{liu2019robertarobustlyoptimizedbert}) to utilizing decoder-only or massive LLM backbones. Models such as the BGE \citep{chen2025m3embeddingmultilingualitymultifunctionalitymultigranularity} series and E5 \citep{wang2024textembeddingsweaklysupervisedcontrastive} established instruction-tuned representations, effectively unifying downstream tasks—like semantic search, clustering, and classification—into a single model via task-specific prefixes. Recognizing the rich semantic understanding capabilities of LLMs, recent research has focused heavily on LLM-augmented training and distillation. The Gecko model \citep{lee2024geckoversatiletextembeddings} demonstrated that lightweight, highly-efficient retrievers can be trained through a two-step distillation pipeline that leverages the vast knowledge of massive LLM teachers. Concurrently, NV-Embed \citep{lee2025nvembedimprovedtechniquestraining} achieved strong performance on the MMTEB leaderboard \citep{muennighoff2023mteb} by transforming decoder-only LLMs into generalist embedders using instruction-tuned contrastive learning and the aggressive integration of synthetic, non-retrieval data. \ours\ \citep{lee2025geminiembeddinggeneralizableembeddings} demonstrated state-of-the-art performance on the MMTEB leaderboard due to utilizing synthetic data and excellent generalization to multilingual tasks through the powerful pre-training of Gemini. 

\paragraph{Evolution of Multimodal Embedders} Early \omnimodal\ embedding paradigms, exemplified by dual-tower models like CLIP \citep{radford2021learning} and ALIGN \citep{jia2021scaling}, were limited by their reliance on narrow contrastive learning objectives over simple image--text pairs. Today, the field is gravitating towards \omnimodal\ architectures capable of mapping text, code, images, structured documents, audio, and video into a single, unified, continuous semantic space. Embedding models are trained by extending existing MLLMs for retrieval via multi-stage contrastive training thereby enabling excellent cross-modal retrieval capabilities. SAIL-Embedding \citep{lin2025sailembeddingtechnicalreportomnimodal} further illustrates this shift by employing a content-aware progressive training methodology mapping \omnimodal\ representations seamlessly into industrial recommendation environments (e.g., sequence-to-item prediction). Similarly, \amazonnova\ \citep{AWS2025novaembeddings} and SigLIP 2 \citep{tschannen2025siglip} have demonstrated strong performance in unifying disparate modalities for cross-modal retrieval workflows. 

\paragraph{Architectural Adaptations for Bidirectional Attention} While causal (autoregressive) LLMs excel in generative tasks, their inherently unidirectional attention mechanism imposes unnecessary limits when generating dense, context-aware embeddings. Several innovative frameworks have emerged to circumvent this limitation. MoCa \citep{chen2025mocamodalityawarecontinualpretraining} directly addresses this by introducing modality-aware continual pre-training, utilizing a joint reconstruction objective that denoises interleaved text and image inputs to force bidirectional context-aware reasoning on top of a causal backbone. Similarly, MM-Embed \citep{lin2025mmembeduniversalmultimodalretrieval} tackles the problem of modality bias through modality-aware hard negative mining, ensuring that embedding models do not disproportionally favor text-to-text resonance at the expense of cross-modal relevance.

\paragraph{Adaptation to Enterprise Use Cases} With enterprise and agentic needs scaling to massive contexts and increasingly focused on documents, modern embedders are required to ingest vast informational payloads efficiently. Models utilize specialized visual-document processing (such as tiled mixtures of vision encoders) to embed complex PDFs, charts, and tables which causes the RAG system's quality to be dependent on various parts of the processing pipeline like chunking strategies etc. 

While these preceding architectures have successfully pushed the boundaries of multi-stage distillation, LLM backbone adaptation, and applications to enterprise use cases, they predominantly address these axes in isolation. \Ourstwo\ unifies these capabilities into a single model that spans a breadth of use cases across which the model can be used out-of-the-box.

%% file: sections/modeling.tex
\section{\beginOmniModal\ Gemini Embedding}

In this section we provide technical details of the \beginOmniModal\ \Ourstwo\ in terms of the model architecture, the objective function, and the training recipe.

\subsection{Model Architecture}

The \Ourstwo\  model is built to create holistic representations of inputs of different modalities and of inputs that combine such modalities.
These representations can be used in diverse downstream tasks including retrieval, clustering, classification, and ranking.
\Ourstwo\  leverages the \omniModal\ and cross-modal power of Gemini to build such representations.
The embedding model is initialized from Gemini and further fine-tuned with task-specific, modality-specific, and cross-modality training.
This allows \Ourstwo\  to build representations on top of the vast knowledge already present in the Gemini parameters.
In this sense, initializing \Ourstwo\  from Gemini can be understood as the ``pre-training" stage of the embedding model.

\Ourstwo\ constructs representations in a manner similar to our previous  \ours\ model \citep{lee2025geminiembeddinggeneralizableembeddings}, but with the important difference that different modalities require different steps to convert the raw format into a sequence of tokens.
In \Ourstwo\  we leverage Gemini to do these types of data and format conversions.
In this way, the model can take as input raw images, video or audio in the formats natively supported by Gemini. %

After tokenization, an input sequence $\mathbf{T}$ of $L$ tokens is processed by $\mathcal M$, a transformer with bidirectional attention initialized from Gemini, producing a sequence of token embeddings $\mathbf{T}_\mathrm{embed} = \mathcal{M}(\mathbf{T}) \in \mathbb{R}^{L \times d_\mathcal{M}}$, where $d_\mathcal{M}$ is the transformer model dimension. To generate a single embedding representing all the information in the input, a pooler $\mathcal{P}$ is applied, $\mathbf{P}_\mathrm{embed} = \mathcal{P}(\mathbf{T}_\mathrm{embed}) \in \mathbb{R}^{d_\mathcal{M}}$.
Prior research~\citep{suganthan2025adaptingdecoder} demonstrated that simple pooling strategies can be effective in model adaptation. Therefore we choose mean pooling, and simply average the token embeddings along the sequence axis. Finally, a randomly initialized linear projection $\mathit{f}$ is applied to scale the embedding to the target dimension, $\mathbf{E} = \mathit{f}(\mathbf{P}_\mathrm{embed}) \in \mathbb{R}^{d}$, where $d$ is the output embedding dimension.

\subsection{Training Objective}
\label{sec:training_objective}

The \omniModal\ nature of \Ourstwo\ requires a multi-task and multi-stage type of training.
This way different modalities can be trained in separate tasks.
We used a multitude of single-modality tasks, \omniModal\ tasks, as well as cross-modal tasks.

Similar to our previous version \citep{lee2025geminiembeddinggeneralizableembeddings}, the \omniModal\ \Ourstwo\ model was trained with a noise-contrastive estimation (NCE) loss with in-batch negatives~\citep{oord2018representation}.
The exact loss differs slightly depending on the task being trained. In general, a training example includes a query $q_i$, a positive target $p_i^+$ and (optionally) a hard negative target $p_{i}^-$.
In text-only training tasks, each example also has a prescribed task string $t$, for example "question answering" or "fact checking", describing the nature of the task.
During training, we randomly drop off the task string $t$ to augment the robustness of the model to different modality inputs where the task strings are not used.
The query and passages are embedded as vectors in $\mathbb R^d$:
\begin{equation}
    \mathbf q_i = f(\meanpool(\mathcal M(t \oplus q_i))),\quad \mathbf p^\pm_i = f(\meanpool(\mathcal M(p^\pm_i))).
\end{equation}
Given a batch of size $B$ the loss applied to these embeddings is as follows:
\begin{equation}
    \mathcal L = \frac 1 B \sum_{i=1}^B \left[ -\log \frac{e^{\pairsim(\mathbf q_i, \mathbf p_i^+)/\tau}}{e^{\pairsim(\mathbf q_i, \mathbf p_i^+)/\tau} + 
    e^{\pairsim(\mathbf q_i, \mathbf p_{i}^-)/\tau} +
    \sum_{j=1}^B \mask(i,j)
    e^{\pairsim(\mathbf q_i, \mathbf p_j^+) / \tau}}\right]
\end{equation}
where $\pairsim(\mathbf x, \mathbf y)= \mathbf x^\top \mathbf y / \lVert \mathbf x \rVert \lVert \mathbf y \rVert$ is cosine similarity, and
\begin{equation}
    \mask(i,j) = \begin{cases}
    0 \quad & \text{if }q_i=q_j \text{ or } p_i^+=p_j^+,\\
    1 \quad & \text{otherwise.}
    \end{cases}
\end{equation}
This masking term is particularly relevant for classification tasks, where the number of targets (labels) is small.
It should be noted that the second term in the denominator is omitted if no hard negatives are provided.

In order to support different dimensions of embeddings with a single model, we adapt the above loss using MRL \citep{kusupati2022matryoshka} into $k$ separate losses across $k$ overlapping sub-dimensions of the embedding dimensions (e.g. multi-loss training with one loss for the first 768 embedding dimensions, another for the first 1,536 dimensions, and so on).
\Ourstwo\ provides $d=3{,}072$ dimensional embeddings, with the MRL support optimized for 768 and 1,536 dimensions.

\subsection{Recipe}

We heavily lean on the multi-task nature of our training setup to let the model learn from each of the different tasks that, as mentioned in section \Cref{sec:training_objective}, contribute in different ways to build the unified embedding space across the different modalities.
We adopt the multi-stage training from previous models like Gecko \citep{lee2024geckoversatiletextembeddings} and \ours\ \citep{lee2025geminiembeddinggeneralizableembeddings} as described below.

\paragraph{Pre-Fine-Tuning (PFT)}
To adapt the parameters in the model from auto-regressive generation to encoding, this stage uses as training a large number of potentially noisy query--target pairs in a multi-task setup.
Further, in this stage we find it beneficial to use large batch sizes which provide more stable gradients, mitigating the impact of the noisy inputs. During this stage, only image, text and code tasks are used in our multi-task setup. The examples from each different task are sampled at pre-specified sampling rates to build training batches of a single task.

\paragraph{Fine-Tuning (FT)} The fine-tuning stage for this model is based on training with a large number of text, code, document, image, audio, and video tasks.
Many, but not all, of the tasks in this fine-tuning include examples that contain query, target, and hard negative target triplets.
For this training stage we found it beneficial to tune batch sizes for each task to improve quality on corresponding evaluations. 
In this stage we also sample examples from one single task to build the training batches.
The alignment between modalities is based on training multiple single-modality batches as well as cross-modality ones.
As in the previous stage, training with all the different tasks and modalities require a multi-task training setup and the sampling rates of each of the different tasks are defined empirically.
Empirically, we found that balancing overall performance across all modalities was sensitive to hyper-parameters like sampling rates and batch sizes in the multi-task setup.

\paragraph{Model Soup} To systematize the combination of different checkpoints and obtain additional generalization performance across the different modalities, we average the parameters obtained from individual fine-tuning runs.
We experimented with different combinations of parameters, including averaging checkpoints from the same training run \citep{izmailov2018averaging}, from different training runs \citep{wortsman2022model}, as well as various weighted averages.

%% file: sections/evaluation.tex
\section{Evaluation}
We rigorously evaluate Gemini Embedding 2 across a comprehensive suite of multimodal and unimodal benchmarks, demonstrating its state-of-the-art capabilities in text, image, video, and audio understanding. Unlike competing models that often rely on brittle, task-specific instructions, Gemini Embedding 2 provides a robust, unified latent space that delivers high performance in zero-shot settings without the need for manual prompt engineering.

\subsection{\beginOmniModal\ Retrieval}
\input{tables/multimodal_table}

We evaluate \Ourstwo\ against other multimodal embedding models — \voyagemultimodal\ \citep{VoyageAI2026multimodal35}, \amazonnova\ \citep{AWS2025novaembeddings}, and Google's legacy model  \legacymme\ \citep{google_cloud_multimodal_embeddings} — across a diverse suite of unimodal, cross-modal and multimodal retrieval benchmarks spanning image, text, and video modalities (see \Cref{tab:multimodal_table}). For unimodal image evaluation, we utilize the Google Universal Embedding Challenge (GUIEC) \citep{araujo2022google} which requires instance-level retrieval over a large-sized index consisting of 200,000 images. We also evaluate cross-modal retrieval quality on image-to-text and text-to-image benchmarks including MSCOCO \citep{chen2015microsoftcococaptionsdata}, Flickr30K \citep{plummer2016flickr30kentitiescollectingregiontophrase}, DOCCI \citep{DOCCI} and TextCaps \citep{TextCaps}. These tasks range from challenging the models on basic image captioning to long captions including spatial reasoning and scene text understanding. We embed the images and texts separately using \Ourstwo\ and then retrieve using cosine similarity between queries and documents over the whole test set. We also evaluate on multimodal embedding capabilities by embedding images and texts together. We do visual question answering as a retrieval evaluation using EncyclopedicVQA \citep{Mensink_2023_ICCV} where we embed the image along with the question to retrieve the correct answer. For text-to-video retrieval, we evaluate on Vatex \citep{wang2020vatexlargescalehighqualitymultilingual}, MSR-VTT \citep{Xu_2016_CVPR}, and YouCook2 \citep{zhou2017automaticlearningproceduresweb} where the video is embedded at 1 FPS up to 32 frames.\

\Ourstwo\ achieves the highest global mean score and leads decisively on unimodal image retrieval, text-to-image, image-to-text,  and text-to-video tasks, with particularly strong results on long-caption benchmarks such as DOCCI and TextCaps. The training mixture shows very good capabilities to generalize to third-party evaluation tasks like Vatex, MSR-VTT, and YouCook2 despite not including any specific in-domain training splits of those datasets. 

On the ViDoRe Benchmark V2 \citep{mace2025vidorebenchmarkv2raising} document retrieval benchmark, as presented in \Cref{tab:multimodal_table} \Ourstwo\ achieves a score of 64.9, delivering competitive performance in a task that demands understanding of page-level visual structure, layout, and embedded text. This places \Ourstwo\ ahead of \amazonnova\ (60.6)  and within close range of \voyagemultimodal\ (65.5). \Ourstwo\ also stands out as one of only two models in this comparison to support the full Video/Audio/Image/Text modality set (alongside \amazonnova), making its document retrieval performance particularly noteworthy given the breadth of tasks it is simultaneously optimized for.

\subsection{MMTEB}

\input{tables/mmteb}

The multilingual benchmark MMTEB \citep{enevoldsen2025mmteb} consists of a large collection of individual evaluation tasks covering 250+ languages and 10 task types: Bitext Mining, Classification, Clustering, Instruction Retrieval, Multilabel Classification, Pair Classification, Reranking, Retrieval, STS, and Summarization. 
\Ourstwo\ overall performance, along with the performance of other \omniModal\ models, is presented in \Cref{tab:mmteb} where we also include the modalities supported by each model.

The MMTEB results demonstrate that \Ourstwo\  outperforms other multimodal models on this text-only benchmark, indicating that its expanded multimodal capabilities do not compromise its performance on purely textual tasks. Relative to our previous text-only \ours\ model, the new \omniModal\ \Ourstwo\ shows stronger performance surpassing the Mean (by task) of 68.32 of our previous model with an equivalent of 69.9.
Moreover, our \omniModal\ \Ourstwo\ sets a new state-of-the-art performance level in task-specific evaluations such as MTEB Code v1 \citep{enevoldsen2025mmteb}, which consists of 12 code retrieval tasks in 15 coding languages, and the Code Information Retrieval benchmark, CoIR \citep{li2024coircomprehensivebenchmarkcode}, which includes 10 of coding retrieval tasks in 9 coding languages.
\Cref{tab:mmteb} also shows that our new \Ourstwo\ model achieves performance that is considerably better in these benchmarks than our previous \ours\ text-only model.
Notably, \Ourstwo\ is also considerably better relative to other text-only models and also better than domain-specific models such as voyage-code-3.

\subsection{MSEB}
\input{tables/mseb}

To rigorously evaluate the auditory capabilities of Gemini Embedding 2, we benchmark the model on the Massive Sound Embedding Benchmark (MSEB) \citep{heigold2026massivesoundembeddingbenchmark}. We focus our evaluation on the retrieval split of MSEB. The model is given a spoken query and the task is to find the most relevant information for the query in a large corpus of text documents.

\subsubsection{Experimental Setup}
A persistent challenge in multimodal retrieval is the bottleneck introduced by standard pipelined approaches, where audio is typically transcribed to text before producing the embeddings. To isolate the impact of our unified multimodal architecture, we juxtapose two distinct input modalities:

\begin{enumerate}
  \item \textbf{Gemini Embedding 2 with ASR}: A cascaded baseline where the raw audio signal is first transcribed into text via an Automatic Speech Recognition (ASR) system, and the resulting text is subsequently encoded.
  \item \textbf{Gemini Embedding 2 with audio}: Our proposed approach, which directly processes raw audio inputs without intermediate textual transcription.
\end{enumerate}

We utilize Mean Reciprocal Rank at 10 (mrr@10) as our principal evaluation metric. The retrieval setup is further stratified into two key partitions to assess generalization: PassageInLang (intra-lingual retrieval within the same language) and PassageCrossLang (cross-lingual retrieval).

\subsubsection{Results}

As shown in \Cref{tab:mseb_results}, the results demonstrate that utilizing native audio processing significantly enhances retrieval performance over the ASR baseline. As shown, Gemini Embedding 2 with native audio achieves an average retrieval mrr@10 of 73.99, yielding a substantial improvement over the ASR-based approach (70.40).

Breaking down the task partitions, we observe consistent gains across varying degrees of linguistic complexity:

\paragraph{PassageInLang:} Direct audio modeling improves same-language retrieval by +2.0 points (75.58 vs. 73.58). The performance gap between the cascade baseline and \Ourstwo\ highlights a structural flaw in pipeline architectures. The cascade system (ASR → Retrieval) in this experiment—suffers heavily from error propagation. If the ASR system misinterprets an ambiguous audio snippet and commits to an incorrect text output, the downstream retrieval system faces a fundamentally altered query, leading to poor search results. \Ourstwo\ overcomes this bottleneck by natively encoding the raw audio directly. Instead of forcing a "hard" textual decision (e.g., "recognize speech" vs. "wreck a nice beach"), the resulting embedding preserves the inherent ambiguity of the original acoustic signal. This robust, continuous representation gives the system a significantly better chance of surfacing the correct retrieval results by preserving rich acoustic cues (e.g., prosody, intonation, and emphasis).

\paragraph{PassageCrossLang:} Notably, the performance delta widens in cross-lingual setups. Native audio embeddings yield a striking +5.01 point enhancement (72.56 vs. 67.55). The dramatic jump in PassageCrossLang validates that the modality-agnostic latent space of Gemini Embedding 2 deeply aligns semantic features regardless of the source audio's spoken language, generalizing robustly beyond the strict phonetic bounds parameterized by an intermediate ASR transcriber.

In aggregate, the MSEB benchmark corroborates that Gemini Embedding 2 successfully models contiguous raw audio, effectively consolidating a holistic representation that significantly outperforms transcription-reliant bottlenecks.

%% file: tables/multimodal_table.tex
\begin{table}[t]
      \centering
      \caption{Comparison of embedding models on retrieval benchmarks. Our model shows strong performance across a variety of unimodal, cross-modal,  and multimodal retrieval tasks. $^\dagger$: Average over intersection of tasks where the metrics are available for all models. Modality abbreviations: V=Video, A=Audio, I=Image, T=Text.
  $^\ddagger$: Reported by accessing available APIs unless self-reported. 
      }
      \label{tab:multimodal_table}
      \resizebox{\columnwidth}{!}{%
      \begin{tabular}{llS[table-format=2.2,table-number-alignment=center]S[table-format=2.2,table-number-alignment=center]|S[table-format=2.2,table-number-alignment=center]S[table-format=2.2,table-number-alignment=center]}
      \toprule
       &  & \textsc{Gemini} & \textsc{Amazon Nova$^\ddagger$} & \textsc{Voyage-3.5- $^\ddagger$} & \textsc{multimodal$^\ddagger$} \\
       &  & \textsc{Embedding 2} & \textsc{MME} & \textsc{multimodal}  & \textsc{embedding@001} \\
       &  &  &  &  & \scriptsize{Legacy Google model} \\
      \midrule
      \multirow{2}{3.5cm}{\textbf{Image $\to$ Image} \\ (Recall@1)}
       & \quad GUIEC \citep{guiec}    & {\textbf{79.4}} & {68.6} & {69.4} & {69.5} \\
       & \quad ImageNet \citep{5206848} & {\textbf{83.6}} & {-}   & {-}   & {71.8}   \\
      \midrule
      \multirow{5}{3.5cm}{\textbf{Text $\to$ Image} \\ (Recall@1)}
       & Mean$^\dagger$           & {\textbf{80.5}} & {71.6} & {75.8} & {69.5} \\
      \noalign{\vskip 4pt}
       & \quad MSCOCO \citep{chen2015microsoftcococaptionsdata}    & {\textbf{62.9}} & {57.2} & {58.1} & {53.1} \\
       & \quad Flickr30k \citep{plummer2016flickr30kentitiescollectingregiontophrase} & {89.1}          & {81.6} & {\textbf{89.9}} & {81.4} \\
       & \quad DOCCI \citep{DOCCI}     & {\textbf{93.4}} & {84.0} & {83.8} & {-} \\
       & \quad TextCaps \citep{TextCaps} & {\textbf{89.6}} & {76.0} & {79.4} & {74.0} \\
      \midrule
      \multirow{5}{3.5cm}{\textbf{Image $\to$ Text} \\ (Recall@1)}
       & Mean$^\dagger$           & {\textbf{91.2}} & {81.6} & {85.9} & {83.4} \\
      \noalign{\vskip 4pt}
       & \quad MSCOCO \citep{chen2015microsoftcococaptionsdata}   & {\textbf{78.8}} & {68.3} & {74.5} & {68.2} \\
       & \quad Flickr30k \citep{plummer2016flickr30kentitiescollectingregiontophrase} & {\textbf{97.4}} & {87.5} & {94.5} & {94.0} \\
       & \quad DOCCI  \citep{DOCCI}   & {\textbf{91.3}} & {76.5} & {77.4} & {-} \\
       & \quad TextCaps \citep{TextCaps}  & {\textbf{97.4}} & {88.9} & {88.6} & {88.1} \\
      \midrule
      \multirow{4}{3.5cm}{\textbf{Text $\to$ Video} \\ (NDCG@10)}
       & Mean           & {\textbf{63.1}} & {54.0} & {49.9} & {49.2} \\
      \noalign{\vskip 4pt}
       & \quad Vatex \citep{wang2020vatexlargescalehighqualitymultilingual}   & {\textbf{68.8}} & {60.3} & {55.2} & {54.9} \\
       & \quad MSR-VTT \citep{Xu2016MSRVTTAL} & {\textbf{68.0}} & {67.0} & {63.0} & {57.9} \\
       & \quad YouCook2 \citep{zhou2017automaticlearningproceduresweb}& {\textbf{52.5}} & {34.7} & {31.4} & {34.9} \\
      \midrule
      \multirow{2}{3.5cm}{\textbf{Image+Text $\to$ Text} \\ (Recall@20)}
       & \quad EncyclopedicVQA \citep{Mensink_2023_ICCV} & {\textbf{71.5}} & {-} & {58.6} & {-} \\
       & \rule{0pt}{14pt} & {~} & & & \\
      \midrule
      \multirow{2}{3.5cm}{\textbf{Document Retrieval} \\ (NDCG@10)}
       & \quad ViDoRe V2 \citep{mace2025vidorebenchmarkv2raising} & {64.9} & {60.6} & {\textbf{65.5}} & {28.9} \\
       & \rule{0pt}{14pt} & {~} & & & \\
      \midrule
      \textbf{Overall Performance$^\dagger$} & & {\textbf{77.2}} & {68.2} & {70.0} & {64.1} \\
      \midrule
      \textbf{Modality} &  & {V/A/I/T} & {V/A/I/T} & {V/I/T} & {I/T} \\
      \bottomrule
      \end{tabular}
      }
      \end{table}

%% file: tables/mmteb.tex
\begin{table}[t]
\centering
\caption{Comparison of multimodal and text-only embedding models on the Massive Text Embedding Benchmark, MTEB(Multilingual), MTEB Code v1, and CoIR benchmarks. Modality abbreviations: V=Video, A=Audio, I=Image, T=Text. $^*$: only self-reported the aggregated MTEB(Multilingual) mean score. $^\dagger$: \textsc{voyage-3.5} for MTEB(Multilingual) and \textsc{voyage-code-3} in CoIR. $^\ddagger$: Results were not reported.}
\label{tab:mmteb}
\resizebox{\columnwidth}{!}{%
\begin{tabular}{llll|ll}
\toprule
 & & \textsc{Gemini} & \textsc{Amazon Nova$^*$} & \textsc{Gemini} & \textsc{voyage-3.5/$^\dagger$} \\
 & & \textsc{Embedding 2}   & \textsc{MME} & \textsc{Embedding} & \textsc{voyage-code-3} \\
\midrule
\multirow{2}{3.5cm}{\textbf{MTEB(Multilingual)} \footnotesize{\citep{enevoldsen2025mmteb}}} & Mean (Task) & \textbf{69.9}  & 63.8  & 68.4 & 58.5 \\
& Mean (Type) & \textbf{61.2} &   &  59.6 & 51.9  \\
\cmidrule{2-6}
 & - Bitext Mining & \textbf{85.4} &    & 79.3 & 60.5   \\
 & - Classification & \textbf{73.1} &    & 71.8 & 58.5  \\
 & - Clustering & \textbf{55.3}  &  & 54.6 & 45.9  \\
 & - Inst. Retrieval & 2.9 &    & 5.2 & \textbf{6.5}  \\
 & - Multilabel Class. & \textbf{32.2} &   & 29.2  &  21.7   \\
 & - Pair Class. & 83.2 &   & \textbf{83.6} & 76.0   \\
 & - Reranking & \textbf{69.0} &   & 65.7  & 64.2   \\
 & - Retrieval & \textbf{70.0} &   & 67.7 & 64.0   \\
 & - STS & \textbf{79.4}&   & \textbf{79.4} & 70.0   \\
\midrule
 \multirow{2}{3.5cm}{\textbf{MTEB(Code)} \footnotesize{\citep{enevoldsen2025mmteb}}} &  \multirow{2}{*}{Mean} & \multirow{2}{*}{\textbf{84.0}} & \multirow{2}{*}{--}$^\ddagger$  & \multirow{2}{*}{76.0} & \multirow{2}{*}{--}$^\ddagger$  \\
 & & & \\
 \midrule
 \multirow{2}{3.5cm}{\textbf{CoIR} \footnotesize{\citep{li2024coircomprehensivebenchmarkcode}}} &  \multirow{2}{*}{Mean} & \multirow{2}{*}{\textbf{82.3}} & \multirow{2}{*}{--}$^\ddagger$  & \multirow{2}{*}{73.9} & \multirow{2}{*}{78.5}  \\
 & & & \\
 \midrule
\textbf{Modality} &  & {V/A/I/T} &  {V/A/I/T} & {T} & {T} \\
\bottomrule
\end{tabular}
}
\end{table}

%% file: tables/mseb.tex
\begin{table}[b]
\centering
\caption{Results on the passage retrieval split of the MSEB benchmark. Utilizing native audio input consistently enhances retrieval performance over the ASR baseline, yielding robust gains in both intra-lingual and cross-lingual generalization setups.}
\label{tab:mseb_results}
\begin{tabular}{@{} l c c c @{}}
\toprule
\multirow{2}{*}{\textbf{Model Setup}} & \textbf{Average} & \multicolumn{2}{c}{\textbf{Retrieval Split (mrr@10)}} \\
\cmidrule(lr){3-4}
 &  & Passage In-Lang & Passage Cross-Lang \\
\midrule
Gemini Embedding 2 w/ ASR & 70.40 & 73.58 & 67.55 \\
Gemini Embedding 2 w/ Native Audio & \textbf{73.99} & \textbf{75.58} & \textbf{72.56} \\
\bottomrule
\end{tabular}
\end{table}

%% file: sections/ablation.tex
\section{Ablation Study}
To better understand how Gemini Embedding 2 achieves great performance across many different tasks and languages, we provide a systematic analysis of our training recipe.

\input{tables/specialized_domain_i2t}

\subsection{Generalization to specialized domains}

To rigorously assess the versatility and multimodal alignment of \Ourstwo\ in specialized contexts, we evaluated its zero-shot image-to-text retrieval capabilities across a diverse suite of domain-specific datasets. To ensure a comprehensive evaluation, we selected datasets corresponding to distinct real-world applications: microscopy and bioscience (MicroVQA~\cite{burgess2025microvqa}), fine art (ArtCap~\cite{lu2022artcap}), astronomy (AstroLLaVA~\cite{zaman2025astrollava}), and culinary arts (Recipe1M~\cite{marin2021recipe1m+}). Formulated as a standard Recall@5 (R@5) benchmark, we compared our model against an array of open-source and proprietary vision-language models (see \Cref{tab:ood_i2t}).

Our findings demonstrate that \Ourstwo\ achieves state-of-the-art performance across all evaluated domains, frequently establishing substantial margins of improvement over existing baselines. For instance, in astronomy (AstroLLaVA) and microscopy (MicroVQA), \Ourstwo\ achieves a R@5 of 64.4 and 79.3, respectively, effectively doubling the performance of these baselines in astronomy, and outperforming them by over 48\% in microscopy. On the Recipe1M dataset, it breaks the 90.0 barrier for retrieving both ingredients (90.2) and instructions (92.1), decisively outperforming the next-best model, SigLIP2-Giant (81.2 and 80.4).

Beyond absolute performance margins, our evaluation highlights a notable difference in cross-domain consistency. While the performance of existing model families often fluctuates significantly depending on the target domain, \Ourstwo\ maintains a robust, general-purpose alignment. As shown in \Cref{tab:ood_i2t}, many baseline architectures exhibit incidental performance peaks and valleys across different specialized domains. For instance, the TIPS \citep{maninis2025tipstextimagepretrainingspatial} model family demonstrates strong alignment in the fine art domain, with TIPS-G14 achieving a R@5 of 65.2 on ArtCap. Yet its performance is comparatively much lower on microscopic biological imagery (20.0 on MicroVQA). Similarly, while the SigLIP2 lineage excels at the Recipe1M dataset (scoring up to 81.2), it struggles to capture the visual semantics of ArtCap (dropping to 8.4). Conversely, \Ourstwo\ does not exhibit these sharp, domain-dependent fluctuations. Instead, it offers a consistently reliable multimodal embedding space that generalizes predictably across a diverse array of highly specialized tasks.

Ultimately, these results underscore the unprecedented robustness of \Ourstwo 's representations out-of-the-box. Users—ranging from bench biologists and astrophysicists to culinary platforms and digital humanities researchers—can readily integrate \Ourstwo\ into their diverse workflows to power highly-accurate, domain-aware, multimodal retrieval systems.

\subsection{Impact of synthetic data}
\vspace{-0.3em}

The text-only \Ours\ model \citep{lee2025geminiembeddinggeneralizableembeddings} showed the effectiveness of the Gemini model to improve the quality of the text data used to train the \Ours\ model.
In this new \Ourstwo\ model, we also used the power of Gemini to improve the quality of the data used to train the model.
We illustrate this with some of the MTEB Code tasks as example of the impact of Gemini when it is used to synthesize high-quality training data.
The results are shown in \Cref{tab:synthetic_data}.
Considering the results of the text-only Gemini Embedding model as baseline, the equivalent results of the \omniModal\ \Ourstwo\ model show some improvement, even before adding any synthetic data. This is remarkable because, as it has been observed in other text-only evaluations, the new \omniModal\ model surpasses the performance of our previous text-only version (refer to \Cref{tab:mmteb} for an MMTEB comparison).
Adding synthetic data generated with Gemini, results in very noticeable improvements in the three MTEB Code tasks subject of this analysis, especially in the CodeFeedbackMT \citep{zheng2024opencodeinterpreterintegratingcodegeneration} task and also in the SyntheticText2SQL and CodeFeedbackST \citep{li2024coircomprehensivebenchmarkcode} ones.
Overall, the use of synthetic data gives a remarkable improvement of +15.81 points in average over our previous \Ours\ model in these challenging code retrieval tasks.

\input{tables/synthetic_data}

\vspace{-1.0em}

\subsection{Impact of Fine-Tuning and Pre-Fine-Tuning}
\vspace{-0.3em}
\input{figures/pft_vs_ft}
We compare the performance of the Pre-Fine-Tuning (PFT) checkpoint and the final Fine-Tuning (FT) checkpoint across various image and video understanding tasks. As shown in \Cref{fig:pft_vs_ft},  FT improves performance over PFT across almost all evaluated benchmarks. The improvements on image tasks, while consistent, are relatively modest. The most significant improvements are concentrated in the video evaluations due to the additional video training data in FT.

\subsection{Impact of In-Domain Video Data}
\vspace{-0.3em}

\begin{table*}[t]
\centering
\scalebox{0.87}{
\begin{tabular}{@{} l c c c c c c @{}}
\toprule
\multirow{2}{*}{\textbf{Model Configuration}} & \multicolumn{2}{c}{\textbf{MSR-VTT}} & \multicolumn{2}{c}{\textbf{YouCook2}} & \multicolumn{2}{c}{\textbf{Vatex}} \\
\cmidrule(lr){2-3} \cmidrule(lr){4-5} \cmidrule(lr){6-7}
& nDCG@10 & \multicolumn{1}{c}{$\Delta$} & nDCG@10 & \multicolumn{1}{c}{$\Delta$} & nDCG@10 & \multicolumn{1}{c}{$\Delta$} \\
\midrule
\textbf{Baseline} \\
Gemini Embedding 2 & 68.2 & \multicolumn{1}{c}{--} & 55.9 & \multicolumn{1}{c}{--} & 69.2 & \multicolumn{1}{c}{--} \\
\midrule
\textbf{Fine-Tuned ($\text{FT}_{\text{mix}}$) Models} \\
\quad + \text{MSR-VTT} data ($\text{FT}_{\text{mix-m}}$) & 75.0 & +6.8 & 56.1 & +0.2 & 71.7 & +2.5 \\
\quad + \text{MSR-VTT} \& \text{Vatex} data ($\text{FT}_{\text{mix-mv}}$) & 76.1 & +7.9 & 55.3 & -0.6 & 79.5 & +10.3 \\
\midrule
\textbf{Model Soups (Gemini Embedding 2 : $\text{FT}_{\text{mix-mv}}$)} \\
\quad Ratio 2:1 ($w_{\text{base}}=2, w_{\text{ft}}=1$) & 71.7 & +3.5 & 56.1 & +0.2 & 74.5 & +5.3 \\
\quad Ratio 1:1 ($w_{\text{base}}=1, w_{\text{ft}}=1$) & 73.7 & +5.5 & 56.8 & +0.9 & 76.8 & +7.6 \\
\bottomrule
\end{tabular}
}
\caption{Summary of video metrics (\text{NDCG@10} in \%) for fine-tuned and souped models. The $\Delta$ columns indicate absolute percentage point differences relative to the Gemini Embedding 2 baseline. Adding targeted data improves in-domain performance but can slightly degrade out-of-domain tasks (e.g., YouCook2 dipping by 0.6\%), whereas model souping effectively balances these task-specific gains with the original model's robustness.}
\label{tab:video_metrics}
\end{table*}

Comparing the fine-tuned models built on top of \Ourstwo, \Cref{tab:video_metrics} shows that the evaluation metrics are highly sensitive to the addition of targeted, in-domain data. Note that we add the in-domain data into the finetuning mixture and train one epoch of the added data. With only a few thousand steps of training and modest O(k) data quantities , we can drive significant improvements in targeted tasks (e.g., adding MSR-VTT and Vatex's training splits pushes MSR-VTT to 76.1\% and Vatex to 79.5\%). However, this narrow focus can lead to slight degradations in out-of-domain tasks (such as YouCook2 dipping to 55.3\%). Interestingly, the newly fine-tuned weights remain highly compatible with the original base model through model souping. Simple interpolation of the souping weights (such as the $2 \times \text{Gemini Embedding 2} + 1 \times \text{fine-tuned}$ or $1 \times \text{Gemini Embedding 2} + 1 \times \text{fine-tuned}$ mixtures) effectively brings back the video performance gains, in several cases yielding better results across the board than the baseline by balancing task-specific knowledge with the robustness of the original model.

%% file: tables/specialized_domain_i2t.tex
\begin{table*}[t]
    \centering
    \caption{Image-to-Text Retrieval (R@5) performance across various Specialized Domains.} 
    \label{tab:ood_i2t}
    \resizebox{\textwidth}{!}{
    \begin{tabular}{ll ccc cc}
        \toprule
        \multirow{2}{*}{\textbf{Model}} & \multirow{2}{*}{\textbf{ Variant}} & \multirow{2}{*}{\textbf{MicroVQA}~\cite{burgess2025microvqa}} & \multirow{2}{*}{\textbf{ArtCap}~\cite{lu2022artcap}} & \multirow{2}{*}{\textbf{AstroLLaVA}~\cite{zaman2025astrollava}} & \multicolumn{2}{c}{\textbf{Recipe1M}~\cite{marin2021recipe1m+}} \\
        \cmidrule(lr){6-7}
        & & & & & \textbf{Ingredients} & \textbf{Instructions} \\
        \midrule
        \multirow{3}{*}{CLIP~\cite{radford2021learning}} & Base Patch32 & 34.1 & 34.1 & 21.2 & 64.6 & 61.1 \\
         & Large Patch14 & 44.4 & 49.4 & 28.8 & 76.5 & 74.6 \\
         & Large Patch14-336 & 46.7 & 52.2 & 31.6 & 76.0 & 75.6 \\
        \midrule
        ALIGN~\cite{jia2021scaling} & Base & 48.1 & 49.2 & 18.4 & 70.3 & 70.8 \\
        \midrule
        \multirow{3}{*}{SigLIP 2~\cite{tschannen2025siglip}} & Base Patch16-256 & 23.0 & 16.3 & 6.3 & 69.8 & 70.7 \\
         & Large Patch16-384 & 27.4 & 7.3 & 11.0 & 78.7 & 78.3 \\
         & Giant Patch16-384 & 33.3 & 8.4 & 13.2 & 81.2 & 80.4 \\
        \midrule
        \multirow{3}{*}{TIPS~\cite{maninis2025tipstextimagepretrainingspatial}} & Base Patch14 & 14.8 & 59.3 & 6.9 & 60.7 & 59.3 \\
         & Large Patch14 & 21.5 & 59.9 & 8.9 & 61.3 & 63.0 \\
         & Giant Patch14 & 20.0 & 65.2 & 10.1 & 66.0 & 65.6 \\
        \midrule
        Voyage-3.5-multimodal & $-$ & 53.3 & 48.7 & 30.3 & $-$ & $-$ \\
        \midrule
        Gemini Embedding 2 & $-$ & \textbf{79.3} & \textbf{67.7} & \textbf{64.4} & \textbf{90.2} & \textbf{92.1} \\
        \bottomrule
    \end{tabular}
    }
\end{table*}

%% file: tables/synthetic_data.tex
\begin{table}[t]
\centering
\caption{Results on selected MTEB Code v1 tasks using synthetic datasets.  Ablation models exclude souping. }
\label{tab:synthetic_data}
\resizebox{\columnwidth}{!}{%
\begin{tabular}{ll|ccc}
\toprule
& \textbf{Average} & {CodeFeedbackMT} & {CodeFeedbackST} & {SyntheticText2SQL}  \\
\midrule
\textbf{Gemini Embedding} & 70.5 & 56.3 & 85.3 & 70.0 \\
\midrule
\textbf{Gemini Embedding 2} \textbf{w/o Synthetic} & 73.0 & 57.9 & 85.5 & 75.7 \\
\textbf{Gemini Embedding 2} \textbf{w/ Synthetic} & \textbf{86.3} (+15.8) & \textbf{92.3} & \textbf{88.6} & \textbf{78.1} \\
\bottomrule
\end{tabular}
}
\end{table}

%% file: figures/pft_vs_ft.tex
\begin{figure}[t]
\centering
\includegraphics[width=1.0\textwidth]{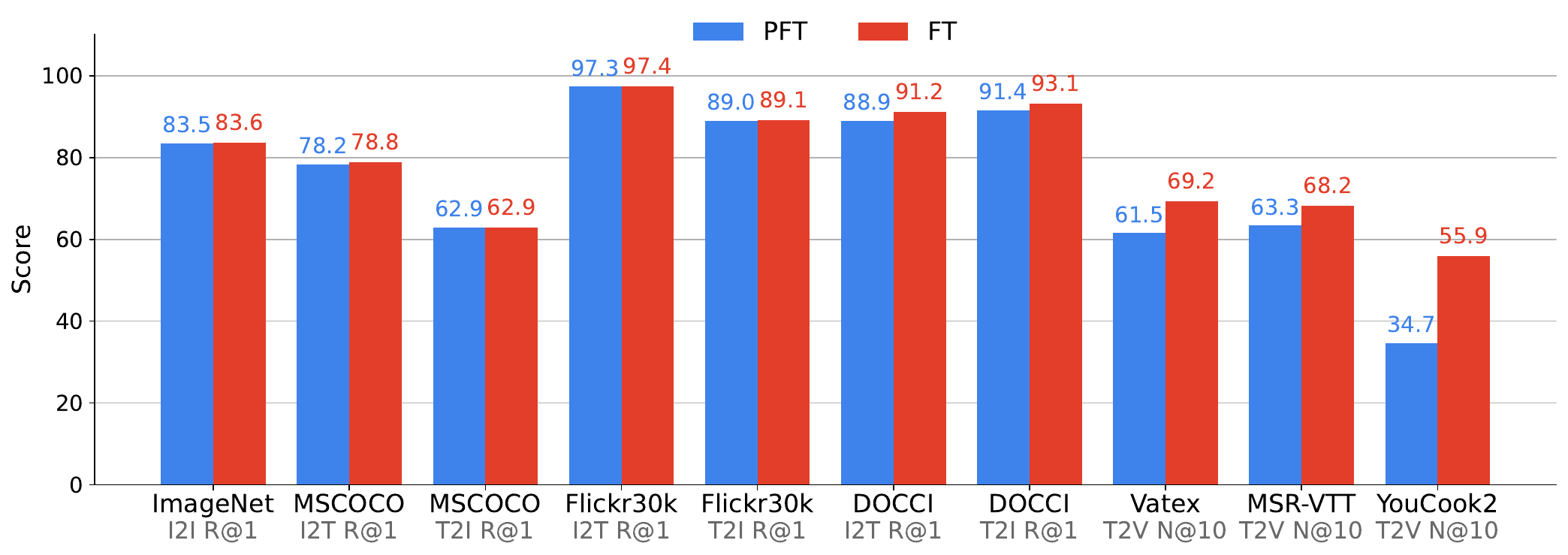}
\caption{Comparing Pre-Fine-Tuning (PFT) and Fine-Tuning (FT) checkpoints on multimodal evals.}
\label{fig:pft_vs_ft}
\end{figure}

%% file: sections/future.tex
\vspace{-1.0em}
\section{Future Work}
The vast native \omniModal\ capabilities of \ourstwo\ unlocks the potential for numerous enterprise use cases like agentic RAG, video recommendation, interleaved multimodal retrieval, etc. without the need for conversion to intermediate modalities. With LLM backbones being highly capable, we believe including other signals from search systems like ranking can be hugely beneficial to improving the retrieval capabilities of embeddings. Agentic RAG use cases also point towards potential future directions of training end-to-end RAG use cases with embeddings being fine-tuned for these enterprise use cases. As the scope of interleaved multimodal applications continues to expand, we invite the broader academic community to contribute novel evaluation frameworks to help benchmark these emerging capabilities.

%% file: sections/conclusion.tex
\section{Conclusion}

\Ourstwo\ represents a transformative step forward in general-purpose representation, delivering a state-of-the-art \omniModal\ successor to our text-only \ours\ model. \Ourstwo\ generalizes well across a wide variety of tasks by seamlessly producing embeddings for arbitrary combinations of interleaved inputs across all modalities including text, image, audio, and video.  By leveraging Gemini’s core multimodal, multilingual and code-centric foundations, the \Ourstwo\ model achieves landmark performance on well-known embedding benchmarks like MSCOCO, Vatex and MMTEB with a particularly significant leap in code retrieval. Our findings highlight its remarkable versatility, showing that it excels not only in general tasks but also across specialized domains such as microscopy, astronomy, and the culinary arts. Furthermore, by demonstrating that native audio input outperforms traditional ASR in retrieval tasks and removing the need for costly task-specific instructions, \Ourstwo\ offers a highly efficient architecture. This unified approach to embedding facilitates a sophisticated cross-data retrieval setup, providing the essential infrastructure for building next-generation agentic systems in tandem with Gemini.

%% file: sections/appendix.tex
\section{Full Results}
\input{tables/mteb_multilingual_full}

\input{tables/mteb_code_full}

%% file: tables/mteb_multilingual_full.tex
\begin{table}[ht]
\centering
\caption{Full results of \ourstwo\ on MTEB (Multilingual).}
\resizebox{0.3365\columnwidth}{!}{%
\begin{tabular}{lc}
\toprule
Task Name & Performance \\
\midrule
AILAStatutes & 49.50 \\
AfriSentiClassification & 59.38 \\
AlloProfClusteringS2S.v2 & 61.75 \\
AlloprofReranking & 84.16 \\
AmazonCounterfactualClassification & 86.99 \\
ArXivHierarchicalClusteringP2P & 63.86 \\
ArXivHierarchicalClusteringS2S & 64.54 \\
ArguAna & 83.60 \\
ArmenianParaphrasePC & 97.56 \\
BUCC.v2 & 99.09 \\
BelebeleRetrieval & 93.81 \\
BibleNLPBitextMining & 34.09 \\
BigPatentClustering.v2 & 41.59 \\
BiorxivClusteringP2P.v2 & 53.10 \\
BornholmBitextMining & 64.14 \\
BrazilianToxicTweetsClassification & 33.21 \\
BulgarianStoreReviewSentimentClassfication & 81.32 \\
CEDRClassification & 57.13 \\
CLSClusteringP2P.v2 & 43.56 \\
CSFDSKMovieReviewSentimentClassification & 54.92 \\
CTKFactsNLI & 87.20 \\
CataloniaTweetClassification & 58.76 \\
Core17InstructionRetrieval & 6.44 \\
CovidRetrieval & 80.14 \\
CyrillicTurkicLangClassification & 95.16 \\
CzechProductReviewSentimentClassification & 68.47 \\
DBpediaClassification & 93.83 \\
DalajClassification & 51.26 \\
DiaBlaBitextMining & 89.00 \\
EstonianValenceClassification & 54.47 \\
FaroeseSTS & 88.83 \\
FilipinoShopeeReviewsClassification & 50.11 \\
FinParaSTS & 32.37 \\
FinancialPhrasebankClassification & 87.16 \\
FloresBitextMining & 90.43 \\
GermanSTSBenchmark & 87.90 \\
GreekLegalCodeClassification & 51.65 \\
GujaratiNewsClassification & 92.19 \\
HALClusteringS2S.v2 & 32.30 \\
HagridRetrieval & 99.19 \\
IN22GenBitextMining & 98.43 \\
IndicCrosslingualSTS & 61.36 \\
IndicGenBenchFloresBitextMining & 99.22 \\
IndicLangClassification & 83.39 \\
IndonesianIdClickbaitClassification & 64.95 \\
IsiZuluNewsClassification & 46.16 \\
ItaCaseholdClassification & 69.68 \\
JSICK & 84.85 \\
KorHateSpeechMLClassification & 26.39 \\
KorSarcasmClassification & 64.39 \\
KurdishSentimentClassification & 87.90 \\
LEMBPasskeyRetrieval & 62.25 \\
LegalBenchCorporateLobbying & 96.37 \\
MIRACLRetrievalHardNegatives & 71.15 \\
MLQARetrieval & 84.51 \\
MacedonianTweetSentimentClassification & 73.13 \\
MalteseNewsClassification & 39.76 \\
MasakhaNEWSClassification & 82.63 \\
MasakhaNEWSClusteringS2S & 60.44 \\
MassiveIntentClassification & 80.86 \\
MedrxivClusteringP2P.v2 & 46.53 \\
MultiEURLEXMultilabelClassification & 4.70 \\
MultiHateClassification & 79.31 \\
NTREXBitextMining & 96.48 \\
NepaliNewsClassification & 97.98 \\
News21InstructionRetrieval & 2.64 \\
\bottomrule
\end{tabular}}
\centering
\resizebox{0.33\columnwidth}{!}{%
\begin{tabular}{lc}
\toprule
Task Name & Performance \\
\midrule
NollySentiBitextMining & 76.46 \\
NordicLangClassification & 90.34 \\
NorwegianCourtsBitextMining & 95.39 \\
NusaParagraphEmotionClassification & 62.17 \\
NusaTranslationBitextMining & 84.47 \\
NusaX-senti & 85.26 \\
NusaXBitextMining & 93.04 \\
OdiaNewsClassification & 95.78 \\
OpusparcusPC & 97.11 \\
PAC & 70.75 \\
PawsXPairClassification & 61.22 \\
PlscClusteringP2P.v2 & 75.65 \\
PoemSentimentClassification & 57.27 \\
PolEmo2.0-OUT & 77.00 \\
PpcPC & 95.40 \\
PunjabiNewsClassification & 83.12 \\
RTE3 & 89.79 \\
Robust04InstructionRetrieval & -0.44 \\
RomaniBibleClustering & 47.88 \\
RuBQReranking & 77.98 \\
SCIDOCS & 25.68 \\
SIB200ClusteringS2S & 43.33 \\
SICK-R & 83.59 \\
SNLHierarchicalClusteringP2P & 59.59 \\
STS12 & 81.07 \\
STS13 & 89.69 \\
STS14 & 85.48 \\
STS15 & 90.67 \\
STS17 & 88.96 \\
STS22.v2 & 70.80 \\
STSB & 85.02 \\
STSBenchmark & 88.68 \\
STSES & 76.66 \\
ScalaClassification & 54.30 \\
SemRel24STS & 74.87 \\
SentimentAnalysisHindi & 74.48 \\
SinhalaNewsClassification & 82.82 \\
SiswatiNewsClassification & 57.63 \\
SlovakMovieReviewSentimentClassification & 93.57 \\
SpartQA & 8.74 \\
SprintDuplicateQuestions & 96.61 \\
StackExchangeClustering.v2 & 92.18 \\
StackOverflowQA & 97.76 \\
StatcanDialogueDatasetRetrieRetrieval & 63.11 \\
SwahiliNewsClassification & 65.71 \\
SwednClusteringP2P & 45.96 \\
SwissJudgementClassification & 61.77 \\
T2Reranking & 67.72 \\
TERRa & 64.52 \\
TRECCOVID & 77.57 \\
Tatoeba & 89.35 \\
TempReasonL1 & 7.77 \\
ToxicConversationsClassification & 85.85 \\
TswanaNewsClassification & 53.92 \\
TweetTopicSingleClassification & 73.15 \\
TwitterHjerneRetrieval & 94.54 \\
TwitterURLCorpus & 88.07 \\
VoyageMMarcoReranking & 71.89 \\
WebLINXCandidatesReranking & 19.01 \\
WikiCitiesClustering & 79.46 \\
WikiClusteringP2P.v2 & 28.51 \\
WikipediaRerankingMultilingual & 93.25 \\
WikipediaRetrievalMultilingual & 94.82 \\
WinoGrande & 69.57 \\
XNLI & 78.95 \\
indonli & 59.21 \\
\bottomrule
\end{tabular}
}
\end{table}

%% file: tables/mteb_code_full.tex
\begin{table}[ht]
\centering
\caption{Full results of \Ourstwo\ on MTEB(Code).}
\begin{tabular}{lc}
\toprule
Task Name & Performance\\
\midrule
AppsRetrieval & 98.60 \\
COIRCodeSearchNetRetrieval & 91.90\\
CodeEditSearchRetrieval & 91.94 \\
CodeFeedbackMT & 92.30 \\
CodeFeedbackST & 88.59 \\
CodeSearchNetCCRetrieval & 96.25 \\
CodeSearchNetRetrieval & 92.96 \\
CodeTransOceanContest & 93.19 \\
CodeTransOceanDL & 33.72 \\
CosQA & 52.05 \\
StackOverflowQA & 97.89\\
SyntheticText2SQL & 78.11 \\
\bottomrule
\end{tabular}
\end{table}

%% file: sections/contributions.tex
\twocolumn[  
    \begin{@twocolumnfalse}
        \section{Contributions and Acknowledgments}
     \end{@twocolumnfalse}
]

\noindent\textbf{Core Contributors} ($^*$: equal contributions)\\
Madhuri Shanbhogue$^*$\\
Zhe Li$^*$\\
Shanfeng Zhang$^*$\\
Gustavo Hern{\'{a}}ndez {\'{A}}brego$^*$\\
Shih-Cheng Huang$^*$\\
Aashi Jain$^*$\\
Daniel Salz\\
Sonam Goenka\\
Chaitra Hegde\\
Ji Ma\\ 
Feiyang Chen\\ 
Jiaxing Wu\\
Tanmaya Dabral\\
Babak Samari\\
Kevin Poulet\\
Daniel Cer\\
Kaifeng Chen\\
Paul Suganathan\\ 
Hui Hui\\
Jovan Andonov\\
Philippe Schlattner\\
Jay Han\\
Iftekhar Naim\\
Wing Lowe\\
Vladimir Pchelin\\
Albert Yang\\
Yi-Ting Chen\\
Zhongli Ding\\
Grace Zhang\\
Georg Heigold\\
Yichang Chen\\
Antoine Reveillon\\
Brendan Mccloskey\\
Wenlei Zhou\\
Dahun Kim\\
Rui Meng\\
Emma Wang\\
Jack Zheng\\
Halley Fede\\
Zhen Yang\\
Keegan Mosley\\
Brian Potetz\\
Sahil Dua\\
Henrique Schechter Vera\\
Shen Gao\\
Hesen Zhang\\
Andreas Hess\\
Hengxuan Ying\\
Alberto Montes\\
Karan Gill\\
Min Choi\\
Sebastian Russo\\
Anja Hauth\\
Jinhyuk Lee\\
Michael Boratko\\
Megan Barnes\\
Vikram Rao\\
Claudiu Musat\\
Cyril Allauzen\\
Ehsan Variani\\
Shankar Kumar\\
Tom Bagby\\
Junyi Jiao\\
Yang Gu\\
Tengxin Li\\
Ayush Agrawal\\
Roberto Santana\\
Dev Nath\\
Stephen Karukas\\
Shuoxuan Han\\
Lucia Loher\\
Alice Twu\\
Nidhi Vyas\\
Siddharth Bhai\\
Frank Palma Gomez\\
Wangyuan Zhang\\
Chaoren Liu\\
Jizheng Yang\\
Steve Qiu\\
Shijie Zhang\\
Sujay Kulkarni\\
Sascha Rothe\\
Sean Nakamoto

\noindent\textbf{Leadership}\\
Raphael Hoffmann\\ 
Zach Gleicher\\ 
Yunhsuan Sung\\ 
Qin Yin\\ 
Tom Duerig\\ 
Mojtaba Seyedhosseini

\clearpage
\onecolumn
\noindent \textbf{Acknowledgement}\\
James Gan, Jon Matthews, Luciano Martins, Patrick Löber, Anna Kelly, Kristen Quan, Roxanne Daniel, Ryan Trostle, Tania Bedrax-Weiss, Srinivasan (Cheenu) Venkatachary, Howard Zhou, Tomas Izo.